\documentclass[letterpaper, 10 pt, journal, twoside]{IEEEtran}

\usepackage{subfig}
\usepackage{amsmath}
\usepackage{mathtools,amssymb}
\usepackage{hyperref}
\usepackage{bm}
\usepackage{color}
\usepackage{xcolor}
\usepackage{cite}
\usepackage{booktabs}
\usepackage{multirow}
\usepackage[linesnumbered,ruled,vlined]{algorithm2e}
\usepackage[binary-units]{siunitx}
\usepackage{makecell}
\usepackage{graphicx}
\usepackage{balance}
\usepackage{algorithmic}
\usepackage{gensymb}
\usepackage{pifont}
\usepackage{tablefootnote}
\usepackage{arydshln}
\usepackage{siunitx}
\usepackage{footnote}
\usepackage{fixltx2e}

\usepackage{eso-pic} 
\usepackage{pdfpages}
\newcommand{\placetextbox}[3]{
  \setbox0=\hbox{#3}
  \AddToShipoutPictureFG*{
    \put(\LenToUnit{#1\paperwidth},\LenToUnit{#2\paperheight}){\vtop{{\null}\makebox[0pt][c]{#3}}}%
  }%
}%

\DeclareMathOperator*{\argmin}{arg\,min}

\usepackage{mwe,tikz}

\newcommand{\robert}[1]{{\color{cyan} Robert: #1}}
\newcommand{\denys}[1]{}
\newcommand{\franta}[1]{}

\newcommand{\changed}[1]{{#1}}

\newcommand{\TODO}[1]{{\leavevmode\newline\noindent\color{purple}\hspace*{0.0cm}\textbf{TODO:} #1}}

\newcommand{\tick}[0]{\ding{52}}
\newcommand{\cross}[0]{\ding{55}}

\newcommand{\disable}[1]{}

\title{
\vspace{0.6em}
Energy-aware Multi-UAV Coverage Mission Planning with Optimal Speed of Flight
}

\author{Denys Datsko, Frantisek Nekovar, Robert Penicka, Martin Saska
\thanks{Manuscript received: August, 15, 2023; Revised October, 14, 2023; Accepted January, 11, 2024.}
\thanks{This paper was recommended for publication by Editor Chao-Bo Yan upon evaluation of the Associate Editor and Reviewers' comments.
}
\thanks{
The authors are with the Multi-robot Systems Group, Faculty of Electrical
Engineering, Czech Technical University in Prague, Czech Republic (\protect\url{http://mrs.felk.cvut.cz/}). 
This work has been supported by the Czech Science Foundation (GAČR) under research project No. 23-06162M, and by the European Union under the project Robotics and Advanced Industrial Production (reg. no. CZ.02.01.01/00/22\_008/0004590), and by CTU grant no. SGS23/177/OHK3/3T/13.%
}
\thanks{Digital Object Identifier (DOI): 10.1109/LRA.2024.3358581.}
}

\renewcommand\subsubsection[1]{\vspace{0pt}\noindent\textbf{#1.}}

\begin{document}
\placetextbox{0.5}{0.956}{
\fbox{
\begin{minipage}{\dimexpr\textwidth-2\fboxsep-2\fboxrule\relax}
D. Datsko, F. Nekovar, R. Penicka and M. Saska, \textbf{Energy-Aware Multi-UAV Coverage Mission Planning With Optimal Speed of Flight}, in IEEE Robotics and Automation Letters, vol. 9, no. 3, pp. 2893-2900, March 2024, \url{https://doi.org/10.1109/LRA.2024.3358581}.
\end{minipage}
}
}%

\bstctlcite{etals}

\setlength{\abovedisplayskip}{6pt}
\setlength{\belowdisplayskip}{6pt}
\setlength{\abovedisplayshortskip}{4pt}
\setlength{\belowdisplayshortskip}{4pt}

\maketitle

\begin{abstract}
This paper tackles the problem of planning minimum-energy coverage paths for multiple UAVs.
The addressed Multi-UAV Coverage Path Planning~(mCPP) is a crucial problem for many UAV applications such as inspection and aerial survey.
However, the typical path-length objective of existing approaches does not directly minimize the energy consumption, nor allows for constraining energy of individual paths by the battery capacity. 
To this end, we propose a novel mCPP method that uses the optimal flight speed for minimizing energy consumption per traveled distance and a simple yet precise energy consumption estimation algorithm that is utilized during the mCPP planning phase.
The method decomposes a given area with boustrophedon decomposition and represents the mCPP as an instance of Multiple Set Traveling Salesman Problem with a minimum energy objective and energy consumption constraint. 
The proposed method is shown to outperform state-of-the-art methods in terms of computational time and energy efficiency of produced paths.
The experimental results show that the accuracy of the energy consumption estimation is on average 97\% compared to real flight consumption.
The feasibility of the proposed method was verified in a real-world coverage experiment with two UAVs.
\end{abstract}

\begin{IEEEkeywords}
Aerial Systems: Applications; Path Planning for Multiple Mobile Robots or Agents; Planning, Scheduling and Coordination
\end{IEEEkeywords}

\vspace{-0.7em}
\section*{Supplementary Material}
{\footnotesize
\vspace{-0.8em}
\noindent \textbf{Video:}  \url{https://youtu.be/S8kjqZp-G-0} 
\\
\noindent \textbf{Code:} \url{https://github.com/ctu-mrs/EnergyAwareMCPP}
\vspace{-0.8em}
}


\section{Introduction}

\IEEEPARstart{T}{he} Coverage Path Planning~(CPP) is frequently needed for deployment of autonomous Unmanned Aerial Vehicles~(UAVs) in applications including agriculture~\cite{santos2020pathPlanningAgriculture}, animal population counting~\cite{Shah2020MultiUAVCoverage}, post-earthquake assessment~\cite{robotics5040026}, disaster management~\cite{Maza2011}, and structure inspection~\cite{Guerrero2013}.
\changed{The goal of the CPP is to find paths that cover the entirety of a given Area Of Interest~(AOI) with the onboard sensor's footprint.
This means covering AOI with the field of view of a camera sensor or, for example, covering the area with sufficiently dense measurements from LiDAR.
}

\begin{figure}[!t]
\includegraphics[width=0.98\columnwidth,trim={0 2cm 0 9.9cm},clip]{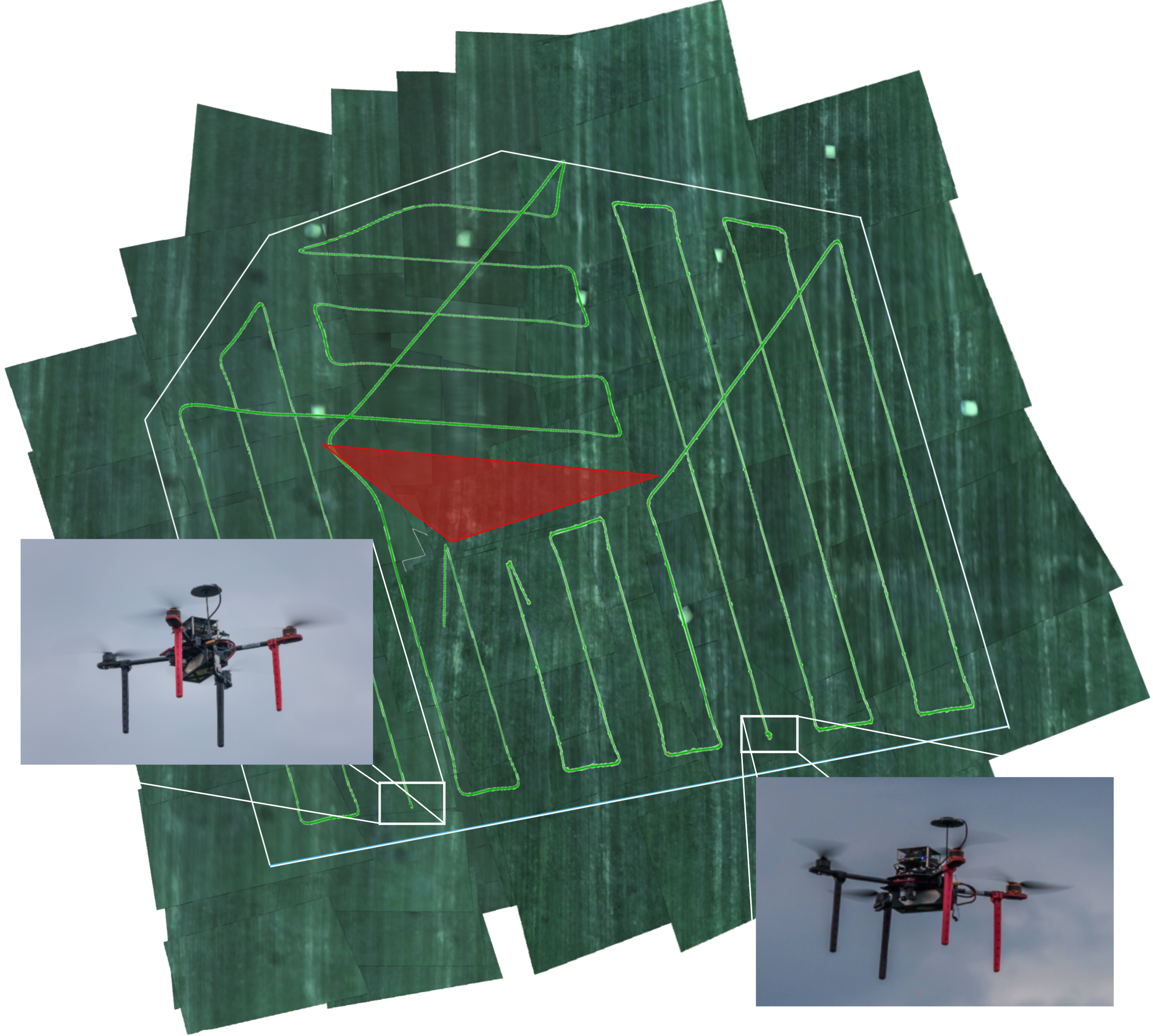} 
\vspace{-0.7em}
\caption{
\label{fig:stitched}
Stitched images taken by two UAVs in a real-world experiment of the proposed energy-aware coverage path planning method together with the photos of used UAVs. 
The white polygon represents the area of interest, the red triangle is a no-fly zone and the green lines are the flown trajectories of the UAVs.}
\vspace{-2.0em}
\end{figure}

Most common optimized objective of CPP with UAVs is the path length~\cite{2019_Cabreira_CPP_UAV_survey}.
However, one of the main limitations of multi-rotor UAVs is the restricted battery capacity, thus limited flight time.
Path length minimization does not directly minimize the energy consumption, as short coverage trajectories with lots of turns require more acceleration, thus energy, than trajectories with mainly straight segments.
Selection of suitable velocity is also important, as low velocities lead to energy wasted to defy gravity without covering much distance, while high velocities leads to increased aerodynamic drag.
Optimizing the energy consumption can lead to shorter overall mission times and/or fewer battery replacements, essential in time-limited applications~\cite{Shah2020MultiUAVCoverage}.
\changed{While few existing approaches minimize energy consumption~\cite{8411478, DiFranco2016}, they only consider single-UAV CPP, do not operate with arbitrary AOI, and require many measurements to estimate the energy consumption for a specific UAV.}
This is due to the complexity of calculating energy consumption, which is a limiting factor for quick evaluation of many possible paths needed for the CPP planning.

\changed{The CPP problem can be represented as the well-known NP-hard Traveling Salesman Problem~(TSP) to minimize a path that visits individual cells of decomposed AOI~\cite{Bahnemann2021GTSP_CPP_boustrophedon}.}
Therefore, solution approximations are necessary to reach reasonable computation times for large areas.
Most recent algorithms~\cite{2019_Cabreira_CPP_UAV_survey} decompose the area into smaller sub-polygons or into a grid pattern. 
Despite that, the CPP is still a challenging problem, especially for the energy-aware CPP with multiple robots. 
Each decomposed sub-polygon has to be covered with a trajectory minimizing energy consumption, which depends on the flight speed and an angle of the back-and-forth coverage trajectory pattern.
\changed{The decomposition has to account for the sub-polygons' coverage energy, as their distribution among multiple UAVs and the order of visit affect the final energy consumption.}
These dependencies between the CPP stages (i.e. the area decomposition, back-and-forth trajectory planning, multi-UAV sub-polygon allocation, and sub-polygons' visit ordering), make the energy-aware Multi-UAV CPP a challenging problem.
\changed{The allocation and ordering problems can be (for a number of trajectories in each sub-polygon) naturally represented as Multiple Set Traveling Salesman Problem~(MS-TSP)~\cite{nekovar2021MSTSP}, which is a multi-robot and budget-constrained extension of the Generalized Traveling Salesman Problem~(GTSP)~\cite{Fischetti1995GTSP}.
}

This letter proposes an energy-aware multi-UAV CPP algorithm that optimizes the energy consumption of vehicle trajectories.
We calculate an optimal flight velocity that maximizes traveled distance~\cite{Bauersfeld2022UAVOptimalSpeed} for a given UAV based on its physical parameters.
\changed{The energy consumption of the multi-UAV CPP solution is minimized by using the optimal velocity for the coverage trajectories, and by using our fast algorithm for estimating trajectory energy during the CPP planning.
The algorithm uses boustrophedon cellular decomposition~\cite{Choset2000}, back-and-forth sweeping patterns in each sub-polygon, the \changed{MS-TSP} representation of the multi-UAV CPP problem and proposed heuristic solver of the MS-TSP.}
By using energy estimation during the planning, we can both minimize the energy consumption and account for the energy consumption constraint imposed by the battery capacity.



\textbf{Contributions} of this letter are as follows.
We propose path energy consumption as an objective in multi-UAV coverage path planning by utilizing optimal flight speed~\cite{Bauersfeld2022UAVOptimalSpeed}.
We introduce a reduction of the CPP problem to MS-TSP formulation.
To the best of our knowledge, this is the first method that directly considers battery capacity constraints during the CPP planning phase to create feasible and energy-efficient multi-UAV missions.
\changed{We verify the method in real-world experiments with multiple UAVs in a visual-coverage experiment depicted in \autoref{fig:stitched}, and we show that the energy consumption estimation of produced trajectories has on average 97\% precision.}
Moreover, our method outperforms the state-of-the-art methods~\cite{Bahnemann2021GTSP_CPP_boustrophedon, Apostolidis2022MultiUAVCoverage,Shah2022LargeScaleCoverage} both in minimizing energy consumption and computational time, although they solve a similar problem without considering optimal speed, energy-aware planning or even no-fly-zones~\cite{Shah2022LargeScaleCoverage}.
Finally, all the methods described here are open-sourced.



   

\section{Related Work}\label{sec:related}


Various methods for solving the CPP have been surveyed in~\cite{Galceran2013CPPsurvey}, and for CPP with UAV applications more recently in~\cite{2019_Cabreira_CPP_UAV_survey}.
They can be classified by different methods of area decomposition.
Some methods use exact cellular decomposition combined with coverage patterns such as sweeping~\cite{Bahnemann2021GTSP_CPP_boustrophedon} or spiral paths~\cite{DiFranco2016}, while others~\cite{Shah2020MultiUAVCoverage} use approximate cellular decomposition and form a graph from the resulting cells.
The utilized boustrophedon cellular decomposition~\cite{Choset2000} offers advantages, as shown in~\cite{Bahnemann2021GTSP_CPP_boustrophedon}, due to usually fewer sub-polygons being decomposed and thus smaller overhead on traveling between them is needed.

The optimization objective of CPP in UAV applications can vary.
Some works optimize total flight path length~\cite{Apostolidis2022MultiUAVCoverage, Shah2020MultiUAVCoverage, Bahnemann2021GTSP_CPP_boustrophedon} as in the Capacited Vehicle Routing Problem~(CVRP)~\cite{ralphs2003capacitatedVRP}, total vehicle flight time~\cite{Bahnemann2021GTSP_CPP_boustrophedon}, number of vehicle path turns~\cite{LI2011876} or the vehicle energy consumed by the UAV moving along the optimized path~\cite{DiFranco2016}.
In long-duration CPP flights, proper energy consumption estimation is needed both to fully utilize the vehicle flight budget and to eliminate mission preemption due to low battery charge.
While some works focus on power consumption in hover conditions, e.g. \cite{Hnidka_2019,Lussier2019enduranceSOTA}, we opted to utilize the method for estimation of energy consumption during a fast flight~\cite{Bauersfeld2022UAVOptimalSpeed} which minimizes the energy per covered distance.

Multi-vehicle generalizations of CPP are the Cooperative Coverage Path Planning~(CCPP)~\cite{mansouri_cooperative_2018} for 3D visual inspection and the multi-vehicle Coverage Path Planning~(mCPP)~\cite{Apostolidis2022MultiUAVCoverage} for area coverage.
The latter utilizes Minimum Spanning Tree~(MST) to approximate cellular decomposition into a square grid together with minimizing the number of turns for energy-efficient paths. 
The algorithm uses Divide Areas based on Robot’s initial Positions (DARP) algorithm~\cite{Kapoutsis2017DARP} for the area decomposition.
A recent work presents a Path Optimization for Population Counting with Overhead Robotic Networks (POPCORN)~\cite{Shah2020MultiUAVCoverage} for mCPP.
The approach uses exact cellular decomposition with problem representation as a series of satisfiability modulo theory instances formulated similarly to~\cite{pathFinding}.
The main objective of the algorithm is to minimize vehicle path lengths. 
The work~\cite{Shah2020MultiUAVCoverage} was extended more recently in~\cite{Shah2022LargeScaleCoverage}, where a similar approach is used, but the original AOI is firstly split to reduce computation times.
The new approach, Split And Link Tiles (SALT)~\cite{Shah2022LargeScaleCoverage}, allows planning for large environments in a reasonable time as it is the number of instances that grows instead of their complexity.
However, splitting the area and coverage of each sub-area separately leads to worse results in terms of path length due to the constrained set of solutions.

A single-UAV approach that utilizes an exact cellular decomposition and back and forth sweeping patterns is presented in~\cite{Bahnemann2021GTSP_CPP_boustrophedon}, where multitude of coverage paths are generated for each sub-cell.
The problem of deciding which paths to select in each sub-cell and in which order to connect them is converted to an instance of a Generalized Traveling Salesman Problem~(GTSP)~\cite{Fischetti1995GTSP}, which is solved exactly.
The objective of the algorithm is to minimize vehicle flight time, assuming UAV stops at each turning point with immediate acceleration change and a trapezoidal speed profile.

A similar approach is also used in~\cite{9743616}, where multi-regional CPP is presented.
The main distinction with the problem presented herein is that regions may be spaced far apart from each other.
The problem is reduced to the Energy constrained Multiple
TSP-CPP~(EMTSP-CPP) and two approaches for solving the problem are presented: Branch and Bound and Genetic Algorithm.
The authors show multiple numerical experiments with two objective functions: minimizing the sum of route lengths and minimizing the maximum route length among all the UAVs.
However, the approach assumes constant velocity coverage, and the energy constraint is only in the form of limited path length. 

In contrast, our method models the UAV up to limited acceleration and uses the optimal velocity to minimize the coverage energy for a given UAV.
Moreover, through accurate and fast energy consumption estimation used during CPP planning, our method can directly use the energy constraints of the battery. A comparison of the discussed approaches is summarized in \autoref{tab:sota_comparison}. 
There, the main difference between "Energy-efficient" and "Energy-aware" fields is that the former means minimizing a metric related to energy consumption (route length, number of turns etc.) while the latter means taking into account energy consumption directly.

\begin{table}[h] 
\vspace{-1em}
\begin{minipage}{\columnwidth}
   \centering 
   \footnotesize 
   {\renewcommand{\tabcolsep}{4.5pt} 
   \caption{Qualitative comparison of the related methods\label{tab:sota_comparison}} 
   \vspace{-0.8em} 
   \resizebox{\columnwidth}{!}{
   \changed {
   \begin{tabular}{ccccccc}
     \toprule 
      & Our & POPCORN+SALT\cite{Shah2022LargeScaleCoverage} & MST\cite{Apostolidis2022MultiUAVCoverage} & GTSP\cite{Bahnemann2021GTSP_CPP_boustrophedon} & Spiral\cite{DiFranco2016} \\
     Multi-UAV & \tick & \tick & \tick & \cross & \cross \\
     Energy-efficient & \tick & \tick & \tick & \tick & \tick \\
     Energy-aware & \tick & \cross & \cross & \cross & \tick \\
     Path length constraint & \cross & \tick & \cross & \cross & \cross\\
     Path energy constraint & \tick & \cross & \cross & \cross & \cross\\
     Arbitrary AOI & \tick & \tick & \tick & \tick & \cross \\
     No-fly-zones & \tick & \cross & \tick & \tick & \cross \\
     
    \bottomrule 
   \end{tabular}
   }
  } 
  }
   \vspace{-1.0em}
   \end{minipage}
\end{table}

\begin{figure*}[t!]
   \centering
   \includegraphics[width=\textwidth]{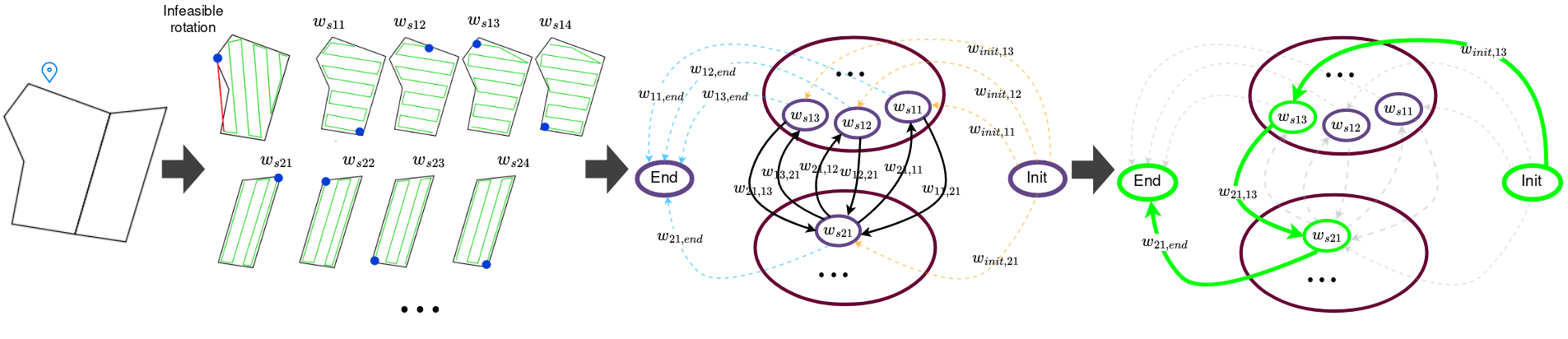}
   \vspace{-2.4em}
   \caption{\label{fig:transformation}
    Problem transformation into an MS-TSP instance. Each value $w_{s\mathbf{x}}$ corresponds to the energy needed to perform sweeping pattern $\mathbf{x}$. Each $w_{x,y}$ denotes the energy needed to get from the last point of sweeping pattern $x$ to the first point of sweeping pattern $y$. Init and End nodes represent UAV's initial and end positions. 
    \vspace{-1.5em}
   }
\end{figure*}

\section{Energy-aware multi-UAV coverage planning\label{sec:method}}
\changed{
We first present the high-level description of the proposed energy-aware multi-UAV CPP algorithm in Sec.~\ref{subsec:cpp_alg}.
Afterward, energy consumption estimation based on trajectory generation and the novel fast energy consumption estimation algorithms are described, both leveraging the pen-and-paper algorithm for calculating the optimal speed per traveled distance from~\cite{Bauersfeld2022UAVOptimalSpeed}.
Finally, each part of the path-planning algorithm is described such as the area decomposition based on Boustrophedon Cellular Decomposition (BCD)~\cite{10.1007/978-1-4471-1273-0_32}, the novel problem conversion to MS-TSP instance, and the MS-TSP solver extending the approach presented in \cite{nekovar2021MSTSP}.
}

\subsection{CPP algorithm\label{subsec:cpp_alg}}
The proposed CPP algorithm is designed to plan paths for a specified amount of UAVs as well as to minimize the number of paths (needed UAV flights) while the maximum path energy stays below a user-defined limit.
For the second case, the planning is done several times, increasing the number of paths generated until the constrained is met.

The method is summarized in \autoref{alg:planning}, where $N_\text{UAV}$ is the number of UAVs, $E_{\text{bound}}$ is the upper bound on the largest path energy, and $N_{min}$ is the minimum number of sub-polygons per UAV after decomposition.
A single planning step for a specified number of paths starts with a greedy calculation of $N_{angles}$ best initial rotation angles for the input polygon (see \autoref{subsec:area_decomposition}).
For each of those angles, an initial AOI is rotated by that angle and decomposed using Boustrophedon Cellular Decomposition (BCD) \cite{10.1007/978-1-4471-1273-0_32} into a set of non-overlapping polygons.
If the number of polygons is smaller than the user-defined limit ($N_{min} \cdot N_{paths}$), polygons with the largest area are divided.
For each of the resulting polygons, different back-and-forth coverage paths with distinct sweeping angle are generated.
Each of these paths is then represented by a node in a weighted graph with a weight equal to the estimated energy consumption of a UAV following that path.
\changed{By grouping all the graph nodes corresponding to the same sub-polygon into a set, we can generate directed edges between each pair of nodes from different sets.
The weight of such edges is equal to the energy needed to move from the last point of the source node path to the first point of the destination node path.}
Finally, initial and end nodes that represent UAVs' initial and end positions are added.
This process is depicted in \autoref{fig:transformation}. 
\setlength{\textfloatsep}{0pt}
 \begin{algorithm}[!t]
 \caption{Energy-aware multi-UAV CPP\label{alg:planning}}
 {\small
 \KwIn{AOI, $N_{\text{UAV}}$, $\text{UAV}\_\text{parameters}$, $E_{\text{bound}}$, $N_{\text{min}}$}
 \KwOut{best\_paths}
    best\_paths = \{\}\\
    \changed{
    $E_{\text{max}} = +\infty$ \hfill $\triangleright$ maximum path energy from best\_paths\\
    $N_{\text{paths}} = N_{\text{UAV}}$ \hfill $\triangleright$ number of paths to generate\\
    }
    \While{$E_{\mathrm{max}} > E_{\mathrm{bound}}$}{
        $E_{\text{max}} = +\infty; E_{\text{tot}} = +\infty$ \\
        get best rotation angles $best\_angles$ \text{ (Sec.~\ref{subsec:area_decomposition})}\\
        \For{$a \in  best\_angles $} {
            sub\_polygons = BCD(rotate(AOI, $a$)) \\
            divide sub\_polygons into at least $N_{\text{min}} \cdot N_{\text{paths}}$ parts \\
            MSTSP\_nodes = \{\} \\
            \For{$polygon \in$ \text{sub\_polygons}}{
                $paths_c$ = get\_coverage\_paths($polygon$, $N_\text{paths}$) \\
                $nodes\_set$ = \{\} \\
                
                \For{$path \in paths_c$}
                {
                    add $path$ to $nodes\_set$
                }
                add $nodes\_set$ to MSTSP\_nodes \\
            }
            construct MSTSP\_instance from MSTSP\_nodes \\
            paths = solve(MSTSP\_instance, UAV\_parameters) \\
            $E = \max_{p \in \text{paths}} \text{ path\_energy}(p)$ \\
            \uIf{$E < E_{\mathrm{max}} $} {
                $E_{\text{max}} = E$ \\
                \changed{
                $E_{\text{tot}} = \sum_{p \in \text{paths}} \text{ path\_energy}(p)$ \hfill $\triangleright$ sum of path energies over generated paths\\
                }
                best\_paths = paths \\
            }
        }
        $N_\text{paths} = \max\{ \text{ceil}(E_{\text{tot}} / E_{\text{bound}}), N_\text{paths} + 1\}$
    }
}

            
\vspace{-0.4em}
\end{algorithm}

After the transformation to MS-TSP, the instance is solved on the created graph representation.
The solver uses Greedy Randomized Adaptive Search Procedure~\cite{Guemri2016} with Greedy Random Search Procedure~GRP  for initial solution generation followed by Tabu Search~(TS) for searching for a better solution.
UAV paths are then recovered from the solution by replacing solution nodes with corresponding coverage paths and edges with a path connecting two adjacent paths.

\subsection{Estimation of path energy consumption}
\changed{The estimation of path energy consumption is based on the pen-and-paper algorithm introduced in \cite{Bauersfeld2022UAVOptimalSpeed}.
Based on the UAV's physical parameters, it can estimate the optimal speed for maximizing the flight distance and power consumption during either flying with such a speed or in hover conditions.
The optimal speed then minimizes the energy consumption of traveling along a straight segment with a fixed distance.}
However, for energy consumption during traveling along a path with turns and speed drops, the model has to be extended.
We use two different energy consumption estimation methods that are different in accuracy and computation time:
\begin{itemize}
    \item estimation by generating a trajectory with Time-Optimal Path Parameterization Based on Reachability Analysis (TOPPRA) \cite{8338417},
    \item \changed{our proposed energy consumption estimation based on path waypoints which does not require trajectory generation and allows real-time planning on large instances.} 
\end{itemize}

\subsubsection{Energy consumption estimation with TOPPRA}
\changed{Even though the TOPPRA~\cite{8338417} is not originally designed for UAV trajectory generation, it works well for our task under certain assumptions.
The energy consumption is estimated based on the UAV trajectory generated for the coverage paths.
In the beginning, the method uses a cubic spline~\cite{mckinley1998cubic} interpolation of the path.
Then, a time parametrization is found using the TOPPRA using the optimal speed as a speed limit.}
The resulting trajectory is sampled and the energy is calculated as the sum of kinetic energy differences between each pair of neighboring points and the energy needed to keep moving with the speed in each waypoint.
The latter is calculated as either the power spent on hover $P_h$ or during flight with the optimal speed $P_r$, both calculated by the pen-and-paper algorithm~\cite{Bauersfeld2022UAVOptimalSpeed}.
For speeds between 0 and the optimal velocity, the power consumption is approximated with the consumption in hover.
This is done due to the complexity of the speed-to-power consumption relation and we show that this approximation works well in real-world experiments described in~\autoref{sec:results}.

\vspace{0.5em}
\subsubsection{Energy consumption estimation without trajectory generation}
As computing a trajectory of long coverage paths is usually a time-consuming task, we introduce a novel energy consumption estimation algorithm that does not perform any trajectory generation.
Accepting $N$ waypoints as an input, the algorithm has a time complexity of $\mathcal{O}(N)$, which allows multiple runs during the path planning phase.

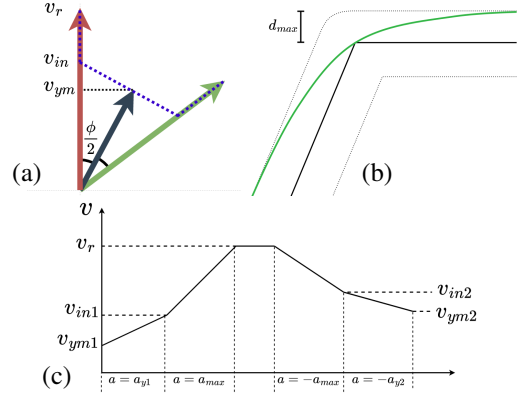
\begin{figure}[t]
  \centering
  \begin{tikzpicture}
  \node[anchor=south west,inner sep=0] (a) at (0,0) {\includegraphics[width=0.32\columnwidth,trim={0 0 0 2.1cm},clip]{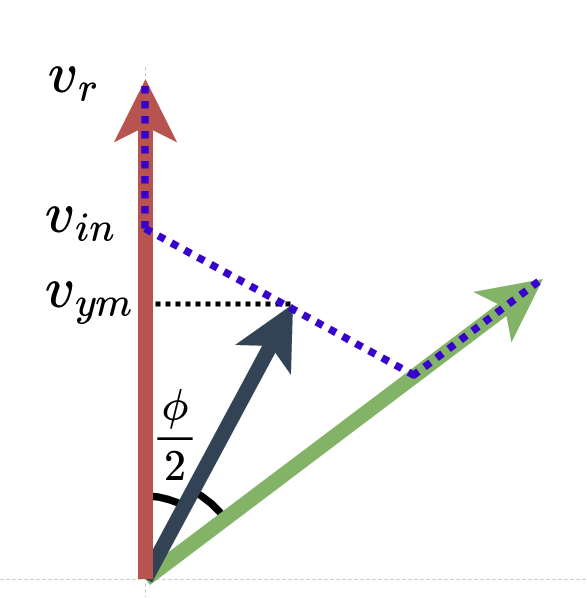}};
  \draw (0.00,0.30) node [text=black, opacity=1] {\normalsize (a)};
  \end{tikzpicture}
  \begin{tikzpicture}
  \node[anchor=south west,inner sep=0] (a) at (0,0) {\includegraphics[width=0.40\columnwidth,trim={0 0 0 2.1cm},clip]{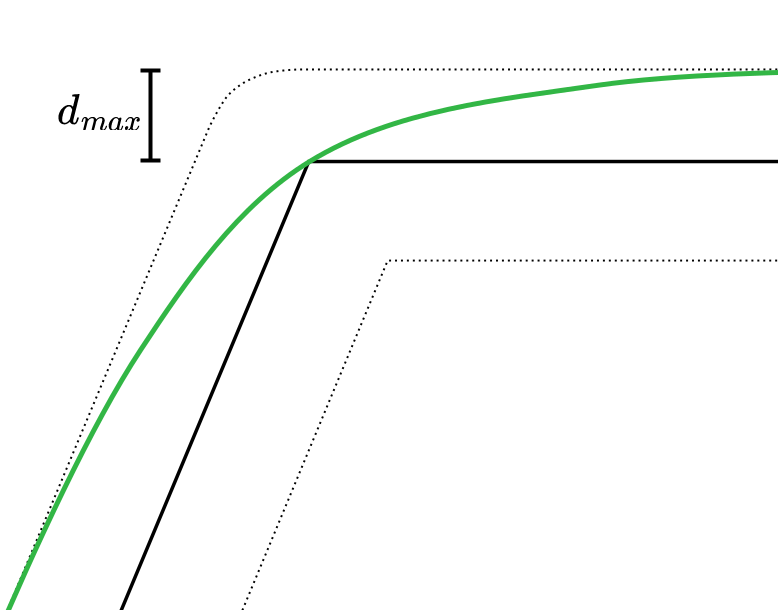}};
  \draw (1.70,0.30) node [text=black, opacity=1] {\normalsize (b)};
  \end{tikzpicture}
    \begin{tikzpicture}
  \node[anchor=south west,inner sep=0] (a) at (0,0) {\includegraphics[width=0.63\columnwidth,trim={0 0 0 2.1cm},clip]{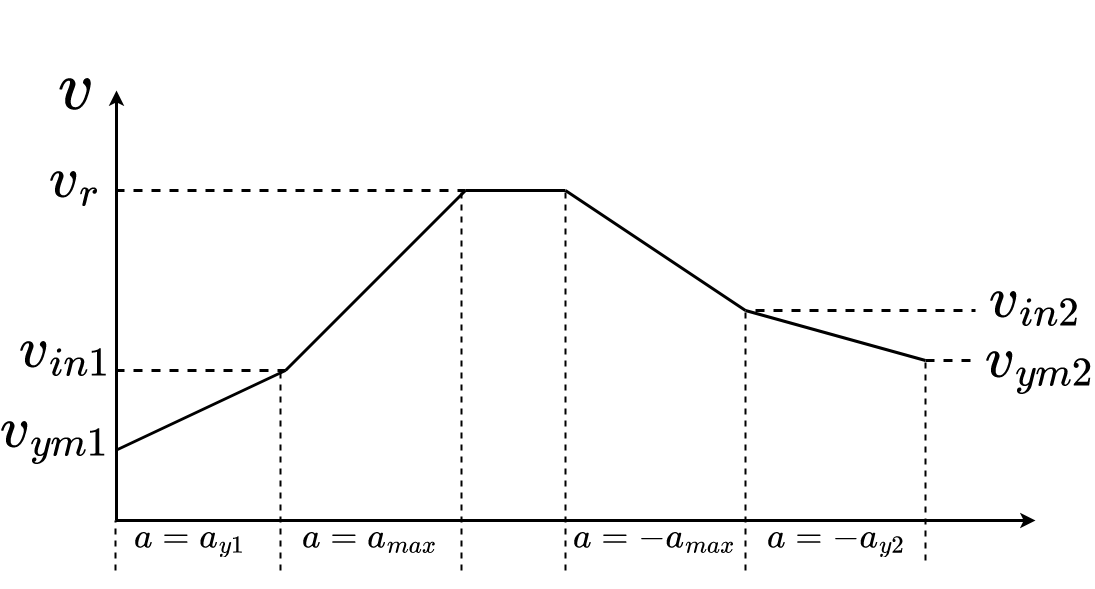}};
  \draw (0.00,0.30) node [text=black, opacity=1] {\normalsize (c)};
  \end{tikzpicture}
  \vspace{-0.5em}
  \caption{
 \label{fig:turn_speeds} (a)~velocity vectors during turn, (b)~assumed turn trajectory, (c)~speed profile projected on path segment.
  }
\end{figure}

Firstly, the turning properties of each turn are calculated considering the situation depicted in \autoref{fig:turn_speeds}.
There, the red vector, located along the $y$ axis, is the speed vector before the turn, the green one is the speed vector after the turn and the gray vector represents the speed in the middle of the turn.
Speed is considered to change along the blue dashed line during the turn.
Value $v_r$ is the optimal speed for the longest range, obtained from the pen-and-paper algorithm, $v_{in}$ is the maximum possible speed before the turn, and $v_{ym}$ is the speed along $y$ axis in the middle of the turn. 
In Fig.~\ref{fig:turn_speeds}(b), the assumed turn trajectory (green line) is shown, together with the original path marked with a solid black line, and the allowed deviation area marked with dotted lines.

The main constraint here is the maximum allowed path deviation, which imposes a constraint on the speed in the middle of the turn (gray vector in Fig.~\ref{fig:turn_speeds}(a)).
If it is large enough, the speed can change from red to green vectors along a line, but in case of smaller deviation allowed, the UAV needs to slow down before turning to not break the constraint.

For each turn, knowing the turning angle $\phi$, values of $v_{in}$ and $v_{ym}$ are calculated as follows:
\begin{equation}
\begin{split}
        &a_x = a_{max} \cdot cos(\phi / 2), \hspace{2em}
        a_y = a_{max} \cdot sin(\phi / 2), \\
        &dv_{x} = min(\sqrt{2 \cdot d_{max} a_x }, \frac{cos(90\degree - \phi) v_r}{2}), \\
        &dv_y = tan(\phi / 2) \cdot dv_x, \\
        &v_{ym} = \frac{dv_x}{tan(\phi / 2)}, \hspace{4.1em}
        v_{in} = v_{ym} + dv_y,
\end{split}
\end{equation}
where $d_{max}$ is the maximal allowed deviation, $a_{max}$ is the acceleration limit, $a_{x}$ and $a_y$ are the accelerations along $x$ and $y$ axes respectively, $dv_x$ and $dv_y$ are differences between red and grey vectors along $x$ and $y$ axes, respectively.

For each path segment between two turns, the speed of the UAV projected on a path segment is assumed to have the profile depicted in Fig.~\ref{fig:turn_speeds}(c).
From this assumption, depending on the specific situation (UAV can reach the optimal speed on the segment, UAV can reach $max(v_{in1}, v_{in2})$, but not $v_r$, etc.), the flight time and energy consumption are found.
For speeds between 0 and $v_r$, the power consumption is approximated with the energy in hover similarly to the estimation with TOPPRA trajectories. 

Here it can be seen that due to considering instant acceleration change and ignoring the exact UAV position during turns that may lead to more complicated trajectories, the algorithm has errors in energy consumption estimation. 
For example, if on a path segment, the UAV can not reach the $v_{ym}$ speed before the turn, it influences the initial speed on the next segment and further ones.
By taking into account each turn and segment separately, these relations are ignored in the proposed algorithm.
Nevertheless, the approximation is fast which allows using it during the planning of CPP, and it is also accurate enough as shown in real-world experiments described in \autoref{sec:results}.

 
 

\subsection{Area decomposition\label{subsec:area_decomposition}}
The first step of the proposed CPP algorithm previously summarized in Sec.~\ref{subsec:cpp_alg} is the area decomposition.
For this purpose, the boustrophedon cellular decomposition (BCD) \cite{10.1007/978-1-4471-1273-0_32} is used.
As the algorithm implies traversing the AOI points from left to right and creating vertical lines, the rotation of the original AOI before decomposition leads to different decomposition results.
For estimation of the best initial rotation, the method described in \cite{Bahnemann2021GTSP_CPP_boustrophedon} is used.
The area is decomposed using each of the boundary segment's rotation as the initial one, and each decomposition is evaluated using the cost function 
\begin{equation}
    w = \sum_{i = 1}^{m} y_{max,i} - y_{min,i},
\end{equation}
where $m$ is the number of sub-polygons, $y_{min,i}$ and $y_{max,i}$ are $y$ coordinates of lowermost and uppermost boundary points of $i$-th segment correspondingly.
Best $N_{angles}$ initial rotations minimizing the cost function are then selected, and the whole path planning algorithm is run for each of them separately.
\changed{The $N_{angles}$ is a user-defined parameter and can vary from 1 to the number of AOI boundary segments.}

After the initial rotation is fixed, the AOI is again decomposed using BCD.
In the output, the number of sub-polygons $m$ should be $ m \geq N_{\text{UAV}} \cdot N_{min}$, where $N_{\text{UAV}}$ is the number of UAVs and $N_{min}$ is the minimum number of sub-polygons per UAV, set by the user.
If and until this condition is not met, the largest sub-polygons are divided into smaller ones.

\changed{The value $N_{min}$ is selected and tuned by the user for each specific planning task.}
Setting it too large results in too many polygons after the decomposition and sub-optimal coverage due to increased time of travel and overlapping coverage between adjacent sub-polygons.
On the other hand, setting it too low can result in uneven work distribution among the UAVs.

\subsection{Sub-polygons coverage}
Having the AOI decomposed into sub-polygons, multiple sweeping coverage paths are generated for each of them.
This algorithm step is based on a statistical analysis showing that the most energy-efficient paths usually have sweeping lines along the longest edge of the polygon.
Therefore, for the $i$-th sub-polygon $4 \cdot min(N_{i,\text{feas}}, N_{e})$ sweeping patterns are generated along $min(N_{i,\text{feas}}, N_{e})$ longest edges in feasible sweeping directions.
The $\changed{N}_{i,\text{feas}}$ is the number of $i$-th feasible polygon sweeping edges, and the $\changed{N_{e}}$ is the maximum number of sweeping rotations in each sub-polygon.

\changed{When creating sweeping paths along the chosen edge, the two possible start points are located at each end of that edge at the distance $\frac{s}{2}$ from it, where $s$ is the sweeping step.
This leads to two corresponding end locations that can also serve as start locations by changing each path's direction, leading to four possible sweeping paths.}
The sweeping direction is feasible if vertical lines along this direction don't leave polygon area (see \autoref{fig:transformation} for an example).

\subsection{Conversion of the Multi-UAV CPP problem to MS-TSP instance}
After having multiple sweeping coverage paths for sub-polygons, each is assigned one graph node, where the weight of the node is equal to the energy needed to fly along it.
All of the nodes associated with same sub-polygon are grouped into the same set.
Edges connecting each pair of nodes from different sets are added with the weight of an edge equal to the energy spent to travel from the last point of the sweeping pattern associated with the source node to the first point of the sweeping pattern associated with the destination node.
This process is depicted in \autoref{fig:transformation}.

\subsection{MS-TSP solver}
Considering that the number of nodes in generated graph may be large and that the algorithm should obtain a solution in a short time just before the flight, we utilize a meta-heuristic algorithm to quickly find possibly only a sub-optimal solution.
This algorithm is a modification of the Greedy Random Adaptive Search Procedure (GRASP) meta-heuristics described in~\cite{nekovar2021MSTSP}.

\begin{figure*}[!htb]
   \centering
   \includegraphics[width=1\textwidth,height=0.25\textwidth]{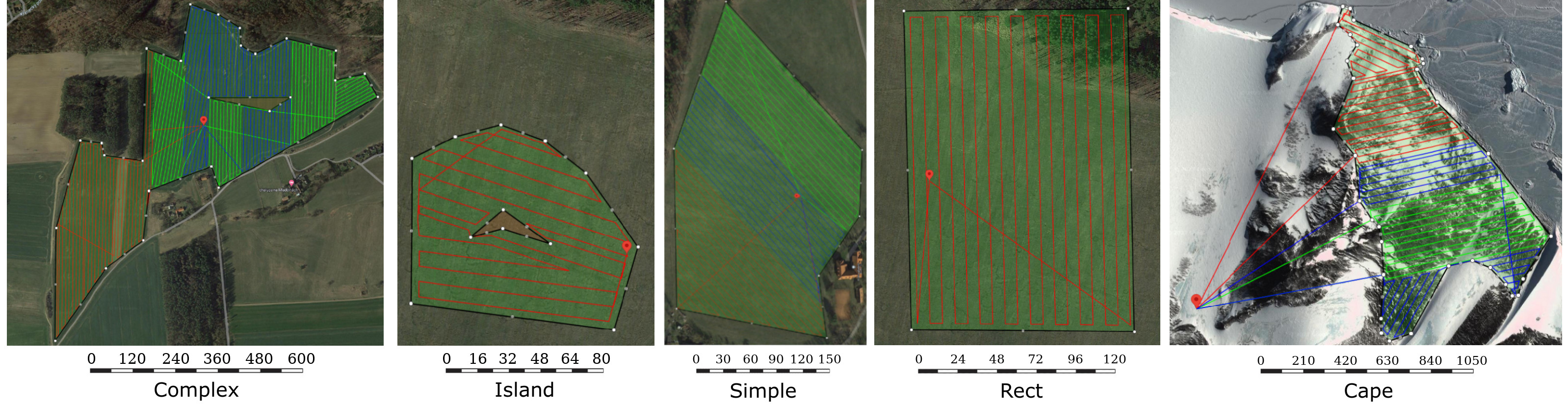}
   \vspace{-2.0em}
   \caption{\label{fig:experiments} 
   Scenarios used for experimental evaluation of the proposed method.
   Shade-of-green polygons represent fly-zones, red polygons are no-fly-zones, and red dot markers represent UAVs' initial positions. 
   All values in the scales are in meters.
   \vspace{-1.5em}
   }
\end{figure*}

Both the total energy consumption and the maximum energy spent by any of the UAVs may be considered as a cost function for MS-TSP.
However, both have their respective drawbacks.
The first can easily lead to uneven workload distribution among the vehicles, while the second function might lead to multiple solutions having the same cost, making the exploitation part of the optimization procedure difficult.
For this reason, both cost functions, referred to as a tuple $(max\_path\_cost, average\_path\_cost)$, are used in tandem.
The maximum energy consumption is used for initial cost comparison, and the average energy consumption is used in the case of two solutions having the same cost. 

 \begin{algorithm}[!ht]
 \caption{MS-TSP solver}
 \label{alg:mstsp}
 \label{algorithm:energy}
 \renewcommand{\algorithmicrequire}{\textbf{Input:}}
 \renewcommand{\algorithmicensure}{\textbf{Output:}}
 {\small
 \KwIn{problem\_instance}
 \KwOut{  instance\_solution}
   $S_{\text{current}}$ = GRASP(problem\_instance)\\
   $S_{\text{best}} = S$\\
   $i = 0$     	\hfill $\triangleright$ iterations without update\\
   $T_{list} = \{S\}$ \hfill $\triangleright$ tabu list\\
  \While{$i < i_{\mathrm{max}}$}{
     $N =$ solution\_neighbourhood($S$)\\
     $S_{\text{current}} = \argmin_{S \in N \setminus T_{\text{list}}}cost(S)$\\
    \If{$cost(S_{\text{current}})< cost(S_{\text{best}})$}{
        $S_{\text{best}} = S_{\text{current}}$\\
    }
   }
  }
  \vspace{-0.4em}
\end{algorithm}

The first step of the MS-TSP solver presented in \autoref{alg:mstsp} is the Greedy Random search Procedure~(GRP) described in detail in~\cite{nekovar2021MSTSP}.
It is used as the initial solution generator where the solution is represented as a list of paths with one path for every UAV.
Each path is a list of graph nodes representing the sweeping patterns and their ordering.
Adaptive Tabu Search (TS) is then used to improve the initial solution using four moves that search through the solution neighborhood.
The solution neighborhood generation uses randomly one of the following four moves:
\begin{enumerate}
    \item \textit{Random shift}: randomly choose a node and move it into a random position among all paths.
    \item \textit{Best shift}: Randomly choose a node. Iteratively find the best position for it and move it there.
    \item \textit{Best swap}: Randomly choose a node. Iteratively find the best node to swap it with and swap them.
    \item \textit{Change direction}: Randomly choose a node and replace it with a random node from the same set.
\end{enumerate}
After applying move procedures 1-3, the moved nodes are replaced with the best (minimizing the cost function) nodes from the same set.

\section{Results\label{sec:results}}
For verification of the proposed method, experiments in simulation and in the real world were performed.
We first compare our approach with state-of-the-art methods in simulation.
Even though our method is the first and only one directly minimizing energy consumption, the other methods minimize related metrics such as the path length, that indirectly minimizes the consumption as well.
The same simulation experiments are also used to show the accuracy of the proposed energy consumption estimation algorithm when compared to the estimation based on the trajectory generation approach.
Second, by conducting a series of real-world experiments, we verify the accuracy of the path energy consumption estimation based on generated trajectory combined with pen-and-paper algorithm for power consumption calculation from~\cite{Bauersfeld2022UAVOptimalSpeed}.

All simulations were done on a computer with an Intel Core i7-10750H CPU with 16 GB of RAM and Ubuntu 20.04 OS. 
A custom build UAV based on Tarot T650\footnote{UAV details - {http://mrs.felk.cvut.cz/research/micro-aerial-vehicles}} frame (see~\autoref{fig:stitched} and \cite{HertJINTHW_paper})  was used for both real-world and simulation experiments.
The calculated optimal speed for the used UAV is \SI{8.39}{\meter/\second}, which is close to the maximum speed at which the UAV with the MRS system~\cite{Baca2021} can fly autonomously.
\changed{It was also used in every simulation experiment as the maximum allowed speed input to the trajectory generation algorithm.}
Values for motor and propeller efficiency (needed for calculating the optimal speed and energy consumption~\cite{Bauersfeld2022UAVOptimalSpeed}) were taken from previous tests of the installed motor and propeller combination (Tarot 4114 320KV motor with Tarot 1555 carbon fiber propeller)\footnote{Motor parameters - https://ctu-mrs.github.io/docs/hardware/motor\_tests.html}.
As a result, for the energy estimation algorithm, the following parameters were used: $v_{r}=\SI{8.39}{\meter / \second}, P_h=\SI{426.03}{\watt}, P_r=\SI{465.23}{\watt} $, calculated with the pen-and-paper algorithm, and $a_{max}~=~\SI{2}{\meter/\second^2}$.
As for other top-level algorithm parameters, $N_{angles} = 7$, and \changed{the tuple $(N_{min}, N_e)$ was set to $(4, 4), (2, 2), (1, 3), (4, 3), (4, 3), (1, 5)$ for scenarios in \autoref{table:3_uavs} in this order.} 


\subsection{Validation in simulation}

\begin{table*}[!htb] 
   \centering 
   \footnotesize 
   {\renewcommand{\tabcolsep}{5.7pt} 
   \caption{Comparison of our method with state-of-the-art for scenarios with three UAVs} 
   \label{table:3_uavs}
   \vspace{-1em} 
\begin{minipage}{\textwidth}
   \begin{tabular}{ccrrrrrrrrrrrr}
     \toprule 
    \multirow{2}{*}{Scenario} & \multirow{2}{*}{$E_b [Wh]$} & \multicolumn{4}{c}{Our} & \multicolumn{4}{c}{POPCORN+SALT \cite{Shah2022LargeScaleCoverage}} & \multicolumn{4}{c}{DARP+MST\cite{Apostolidis2022MultiUAVCoverage} }  \\
    \cmidrule(lr){3-6} \cmidrule(lr){7-10} \cmidrule(lr){11-14}
      & &  $E_{o} [Wh]$ & $E_{t} [Wh]$ & $l [km]$ & $t_c [s]$ & $E_{o} [Wh]$ & $E_{t} [Wh]$ & $l [km]$ & $t_c [s]$ & $E_{o} [Wh]$ & $E_{t} [Wh]$ & $l [km]$ & $t_c [s]$ \\
    \midrule

\multirow{1}{*}{Cape} & $+\infty$ & \textbf{202.2} & 206.6 & 36.1 & \textbf{11.3} & 298.5 & 284.7 & 46.8 & 50.2 & 209.7 & \textbf{200.0} & 34.7 & 33.9  \\
\multirow{1}{*}{Island} & $+\infty$ & \textbf{12.7} & \textbf{11.8} & 1.6 & \textbf{0.4} & - & - & - & - & 13.3 & 12.2 & 1.5 & 5.4  \\
\multirow{1}{*}{Rectangle} & $+\infty$ & \textbf{27.8} & \textbf{25.6} & 4.4 & \textbf{0.1} & 43.0 & 45.4 & 3.9 & 25.6 & 29.0 & 28.1 & 3.8 & 4.9  \\
\multirow{1}{*}{Complex 15} & $+\infty$ & 142.3 & 138.0 & 23.6 & \textbf{6.2} & - & - & - & - & \textbf{133.9} & \textbf{131.0} & 20.1 & 32.6  \\
\multirow{1}{*}{Complex 10} & $+\infty$ & 205.0 & 208.0 & 34.7 & \textbf{7.8} & - & - & - & - & \textbf{195.9} & \textbf{189.1} & 29.8 & 70.3  \\
\multirow{1}{*}{Simple} & $+\infty$ & \textbf{65.1} & \textbf{62.5} & 10.5 & \textbf{0.0} & 132.5 & 125.4 & 11.1 & 170.7 & 72.7 & 68.5 & 9.3 & 12.2  \\

\multirow{1}{*}{Complex 10} & $130$ & \textbf{109.0} & \textbf{103.0} & 36.4 & \textbf{12.8} & - & - & - & - & - & - & - & -  \\
    \bottomrule 
   \end{tabular} 
   \vspace{-1em}
   \end{minipage}
      } 
   \vspace{-0.8em}
\end{table*} 


\begin{table}[!htb] 
   \centering 
   \footnotesize 
   {\renewcommand{\tabcolsep}{4.5pt} 
   \caption{Comparison of our method with state-of-the-art for scenarios with one UAV} 
   \label{table:1_uav}
   \vspace{-1.0em} 
\resizebox{\columnwidth}{!}{
   
   \begin{tabular}{crrrrrrrr}
     \toprule 
    \multirow{2}{*}{Scenario} & \multicolumn{4}{c}{Our} & \multicolumn{4}{c}{GTSP\cite{Bahnemann2021GTSP_CPP_boustrophedon} }  \\
    \cmidrule(lr){2-5} \cmidrule(lr){6-9}
      & $E_{o} [Wh]$ & $E_{t} [Wh]$ & $l [km]$ & $t_c [s]$ & $E_{o} [Wh]$ & $E_{t} [Wh]$ & $l [km]$ & $t_c [s]$ \\
    \midrule 
    \multirow{1}{*}{Cape} & \textbf{479.6} & \textbf{477.3} & 29.2 & \textbf{0.0} & 621.9 & 628.0 & 36.7 & 2.2  \\
\multirow{1}{*}{Island} & \textbf{31.9} & \textbf{30.8} & 1.4 & \textbf{0.1} & 36.5 & 35.3 & 1.6 & 0.4  \\
\multirow{1}{*}{Rectangle} & 71.9 & 65.4 & 3.9 & \textbf{0.0} & \textbf{67.5} & \textbf{63.2} & 3.7 & 0.1  \\
\multirow{1}{*}{Complex 15} & \textbf{381.3} & \textbf{379.0} & 21.7 & 2.2 & 449.6 & 461.1 & 25.7 & \textbf{2.1}  \\
\multirow{1}{*}{Complex 10} & \textbf{554.0} & \textbf{544.7} & 31.7 & \textbf{1.8} & 621.3 & 615.0 & 36.4 & 2.5  \\
\multirow{1}{*}{Simple} & 171.5 & 171.9 & 9.8 & \textbf{0.0} & \textbf{167.6} & \textbf{156.0} & 9.7 & 0.4  \\
    \bottomrule 
   \end{tabular} 
      } 
      }
      \vspace{-1.2em}
\end{table} 

\begin{table}[!htb] 
   \centering 
   \footnotesize 
   {\renewcommand{\tabcolsep}{4.5pt} 
   \caption{Comparison of real and estimated energy consumptions in real-world flights} 
   \vspace{-1em} 
   \label{table:real_world}
   \begin{tabular}{crrrrrr}
     \toprule 
     Experiment & $v_{max}[m/s]$ & $t[s]$ & $E_\text{real}[Wh]$ & $E_{\text{est}}[Wh]$ & $acc [\%]$\\
     \midrule
     
     1 & 2 m/s & 132.0 & 14.65 & 15.06 & 97.27 \\
     2 & 4 m/s & 82.3 & 9.40 & 9.42 & 99.78 \\
     3 & 8 m/s & 58.0 & 6.60 & 6.89 & 95.79 \\
     4 & 10 m/s & 41.8 & 4.80 & 5.19 & 92.48 \\
     5 & 2 m/s & 204.0 & 23.14 & 23.27 & 99.44 \\
     6 & 4 m/s & 107.0 & 12.25 & 12.26 & 99.90 \\
     7 & 8 m/s & 70.8 & 8.01 & 8.36 & 95.63 \\
    \bottomrule 
   \end{tabular}
  } 
   \vspace{-0.0em}
\end{table}

The performance of our method compared to the existing approaches was validated in simulations, together with the precision of the energy consumption estimation. 
Three existing implementations of the state-of-the-art approaches were used for comparison: POPCORN+SALT algorithms from~\cite{Shah2022LargeScaleCoverage} implemented as a Python library; GTSP approach described in~\cite{Bahnemann2021GTSP_CPP_boustrophedon} and implemented as a Robotic Operating System (ROS) node; and DARP+MST solution approach~\cite{Apostolidis2022MultiUAVCoverage} implemented in Java.
Each algorithm was used to generate paths for each of the test scenarios:

\begin{itemize}
    \item{Cape}: one of AOI used in~\cite{Shah2020MultiUAVCoverage}, Cape Crozier on Ross Island in Antarctica, with the same parameters as in~\cite{Shah2020MultiUAVCoverage},
    \item{Island}: simple shape with one No-Fly-Zone~(NFZ) in the middle, 8m coverage footprint width,
    \item{Rect}: a rectangle with no NFZ, 8m coverage footprint width,
    \item{Complex 15}: complex shape with a NFZ, 15m coverage footprint width,
    \item{Complex 10}: same shape as in the previous one, but 10m coverage footprint width,
    \item{Simple}: simple shape with no NFZ, 6m coverage footprint width.
\end{itemize}
The graphical overview of the test scenarios together with an example of planned paths using our method is shown in~\autoref{fig:experiments} with a scale in meters to give a size reference.

The simulation results are shown in \autoref{table:3_uavs} for scenarios with three UAVs and \autoref{table:1_uav} for one UAV.
In the tables, $E_{b}$ is the energy budget of each UAV, $E_o$ is the maximum energy consumption among all the produced trajectories estimated by our fast method without trajectory generation, and $E_{t}$ is the maximum energy consumption estimated with the method based on trajectory generation.
Variable $l$ denotes the produced path length and $t_c$ is the computation time.
Some measurements for the SALT+POPCORN~\cite{Shah2022LargeScaleCoverage} algorithm are skipped (marked `-') as it is not possible to specify no-fly-zones inside the AOI in the provided implementation.
The energy consumption was calculated with the same UAV physical parameters as in the real-world experiments.

It can be seen from \autoref{table:3_uavs} and \autoref{table:1_uav} that the proposed fast energy consumption estimation does not deviate from the method that uses trajectory generation by more than 10\% in all the cases.
\changed{Moreover, for the CPP with one UAV (see~\autoref{table:1_uav}), our method outperforms the GTSP~\cite{Bahnemann2021GTSP_CPP_boustrophedon} quite significantly in energy consumption in four out of six test cases while having very similar energy consumption in simple scenarios, which can be mostly explained by the difference in implementation of algorithm parts}.
At the same time, our method is faster in computation in five out of six cases.
The better performance of our method compared to the similar GTSP~\cite{Bahnemann2021GTSP_CPP_boustrophedon} is caused by minimizing the energy consumption directly inside the algorithm instead of only after the CPP paths are used to create flight trajectories.

Similarly, our algorithm with three UAVs (see \autoref{table:3_uavs}) is better than the DARP+MST~\cite{Apostolidis2022MultiUAVCoverage} method in terms of energy consumption in four cases, while still being close in two others.
\changed{This shows our method's limitation in effectively working with complex shapes.
When the area has many complex structures, the area decomposition may produce many small sub polygons, and during coverage, some areas on the edges of those will be covered twice.
The DARP algorithm ensures that the number of sub areas after decomposition equals the number of UAVs and that each has a point close to the start.
}
Compared to POPCORN+SALT~\cite{Shah2022LargeScaleCoverage}, the proposed method is more energy-efficient in each of the three valid cases.
\changed{One of the advantages of our method is lower computation time in all test cases.
This is partially due to the cellular decomposition approach, where the number of sub-polygons does not necessarily grow with an increased AOI size or decreased sweeping step.
However, it should also be considered that the open-source implementations are written using different programming languages and frameworks.
}
With the energy constraint applied to one of the scenarios (last row of \autoref{table:3_uavs}), our approach is the only one that can adaptively adjust the number of produced paths (six in this case) to satisfy it.

\subsection{Real world experiments}
We performed two types of real-world experiments, first to verify the energy estimation accuracy, and second, to test the feasibility of the method in real deployment with multiple UAVs.
During the experiments, all CPP trajectories were smoothed using the polynomial method described in~\cite{burri2015real-time}.
After receiving trajectories, the flight was fully autonomous, controlled by the MRS UAV system~\cite{Baca2021, MRS2022ICUAS_HW, HertJINTHW_paper} using linear Model Predictive Control \cite{8594266}, and running on the onboard computer.

The precision of the energy estimation compared to the real consumption was verified in a series of seven experiments using coverage paths with different flight speeds.
In each case, a coverage path for a simple polygon was produced. 
While the polygon for coverage was slightly different for experiments number 1-3 and the path was significantly shorter in experiment number 4, in experiments 5, 6 and 7 UAV followed the same path.
The results of these experiments are shown in \autoref{table:real_world}.
There, $v_{max}$ is the maximum speed bound, $t$ is the time of following the trajectory, $E_{real}$ is the measured energy consumption using calibrated flight controller internal circuit, $E_{est}$ is the estimated energy consumption, and $acc$ is the estimation accuracy calculated as ${acc = 100 \cdot(1 - \frac{|E_{real} - E_{est}|}{E_{real}})}$.
The value of $E_{est}$ was estimated by sampling the UAV trajectory produced by the polynomial method~\cite{richter2016polynomial} and running the algorithm described above.
As can be seen from the results, the average estimation accuracy was $97.18\%$, which we consider to be accurate enough for real UAV deployment.

Finally, we test the feasibility of the proposed coverage planning in real deployment with two UAVs.
The overview of the AOI, generated paths, and stitched images taken from cameras mounted on UAVs is shown in \autoref{fig:stitched}.
Recording of the experiment together with the description of our method can be found in the linked video {\footnotesize \url{https://youtu.be/S8kjqZp-G-0}}.





\section{Conclusions\label{sec:conclusion}}
We proposed a new energy-aware multi-UAV coverage path planning algorithm based on exact cellular decomposition and MS-TSP formulation.
This is, to the best of our knowledge, the first solution that directly minimize the energy consumption in the multi-UAV CPP.
By introducing energy awareness during the coverage path planning phase and by using optimal flight speed for coverage, the proposed method can both minimize the energy consumption of the coverage paths and constrain path energy consumption of a single path to allow realistic missions with a limited battery capacity. 
We compared the approach to the state-of-the-art methods on a set of simulation experiments, where it outperformed them in a majority of cases in terms of energy consumption with average energy savings from $0.4\%$ up to $40.4\%$.
Our method also showed better computational times in all but one test scenario.
The proposed energy consumption estimation was tested in real flights, where in each test its accuracy was on average $97\%$.
Moreover, the feasibility of the proposed method was verified in a real coverage experiment with two UAVs.

\balance

\bibliographystyle{IEEEtran}
\bibliography{main}

\begin{thebibliography}{10}
\providecommand{\url}[1]{#1}
\csname url@samestyle\endcsname
\providecommand{\newblock}{\relax}
\providecommand{\bibinfo}[2]{#2}
\providecommand{\BIBentrySTDinterwordspacing}{\spaceskip=0pt\relax}
\providecommand{\BIBentryALTinterwordstretchfactor}{4}
\providecommand{\BIBentryALTinterwordspacing}{\spaceskip=\fontdimen2\font plus
\BIBentryALTinterwordstretchfactor\fontdimen3\font minus
  \fontdimen4\font\relax}
\providecommand{\BIBforeignlanguage}[2]{{%
\expandafter\ifx\csname l@#1\endcsname\relax
\typeout{** WARNING: IEEEtran.bst: No hyphenation pattern has been}%
\typeout{** loaded for the language `#1'. Using the pattern for}%
\typeout{** the default language instead.}%
\else
\language=\csname l@#1\endcsname
\fi
#2}}
\providecommand{\BIBdecl}{\relax}
\BIBdecl

\bibitem{santos2020pathPlanningAgriculture}
L.~C. Santos, F.~N. Santos, E.~J. Solteiro~Pires \emph{et~al.}, ``Path planning
  for ground robots in agriculture: a short review,'' in \emph{IEEE
  International Conference on Autonomous Robot Systems and Competitions
  (ICARSC)}, 2020, pp. 61--66.

\bibitem{Shah2020MultiUAVCoverage}
K.~Shah, G.~Ballard, A.~Schmidt \emph{et~al.}, ``Multidrone aerial surveys of
  penguin colonies in antarctica,'' \emph{Science Robotics}, vol.~5, p. 3000,
  10 2020.

\bibitem{robotics5040026}
A.~Nedjati, G.~Izbirak, B.~Vizvari \emph{et~al.}, ``Complete coverage path
  planning for a multi-uav response system in post-earthquake assessment,''
  \emph{Robotics}, vol.~5, no.~4, 2016.

\bibitem{Maza2011}
I.~Maza, F.~Caballero, J.~Capit{\'a}n \emph{et~al.}, ``Experimental results in
  multi-uav coordination for disaster management and civil security
  applications,'' \emph{Journal of Intelligent {\&} Robotic Systems}, vol.~61,
  no.~1, pp. 563--585, 2011.

\bibitem{Guerrero2013}
J.~A. Guerrero and Y.~Bestaoui, ``Uav path planning for structure inspection in
  windy environments,'' \emph{Journal of Intelligent {\&} Robotic Systems},
  vol.~69, no.~1, pp. 297--311, Jan 2013.

\bibitem{2019_Cabreira_CPP_UAV_survey}
T.~M. Cabreira, L.~B. Brisolara, and P.~R. Ferreira~Jr., ``Survey on coverage
  path planning with unmanned aerial vehicles,'' \emph{Drones}, vol.~3, no.~1,
  2019.

\bibitem{8411478}
T.~M. Cabreira, C.~D. Franco, P.~R. Ferreira \emph{et~al.}, ``Energy-aware
  spiral coverage path planning for uav photogrammetric applications,''
  \emph{IEEE Robotics and Automation Letters}, vol.~3, no.~4, pp. 3662--3668,
  2018.

\bibitem{DiFranco2016}
C.~Di~Franco and G.~Buttazzo, ``Coverage path planning for uavs photogrammetry
  with energy and resolution constraints,'' \emph{Journal of Intelligent {\&}
  Robotic Systems}, vol.~83, no.~3, pp. 445--462, 2016.

\bibitem{Bahnemann2021GTSP_CPP_boustrophedon}
R.~B\"{a}hnemann, N.~Lawrance, J.~J. Chung \emph{et~al.}, ``Revisiting
  boustrophedon coverage path planning as a generalized traveling salesman
  problem,'' in \emph{Proceedings of the 12th Conference on Field and Service
  Robotics}, 2019, pp. 1--14.

\bibitem{nekovar2021MSTSP}
F.~Nekovář, J.~Faigl, and M.~Saska, ``Multi-tour set traveling salesman
  problem in planning power transmission line inspection,'' \emph{IEEE Robotics
  and Automation Letters}, vol.~6, no.~4, pp. 6196--6203, 2021.

\bibitem{Fischetti1995GTSP}
M.~Fischetti, J.~J.~S. González, and P.~Toth, ``The symmetric generalized
  traveling salesman polytope,'' \emph{Networks}, vol.~26, no.~2, pp. 113--123,
  1995.

\bibitem{Bauersfeld2022UAVOptimalSpeed}
L.~Bauersfeld and D.~Scaramuzza, ``Range, endurance, and optimal speed
  estimates for multicopters,'' \emph{IEEE Robotics and Automation Letters},
  vol.~7, no.~2, pp. 2953--2960, 2022.

\bibitem{Choset2000}
H.~Choset, ``Coverage of known spaces: The boustrophedon cellular
  decomposition,'' \emph{Autonomous Robots}, vol.~9, no.~3, pp. 247--253, 2000.

\bibitem{Apostolidis2022MultiUAVCoverage}
S.~Apostolidis, P.~Kapoutsis, A.~Kapoutsis \emph{et~al.}, ``Cooperative
  multi-uav coverage mission planning platform for remote sensing
  applications,'' \emph{Autonomous Robots}, vol.~46, pp. 1--28, 2022.

\bibitem{Shah2022LargeScaleCoverage}
K.~Shah, A.~Schmidt, G.~Ballard \emph{et~al.}, ``Large scale aerial multi-robot
  coverage path planning,'' \emph{Field Robotics}, vol.~2, pp. 1971--1998,
  2022.

\bibitem{Galceran2013CPPsurvey}
E.~Galceran and M.~Carreras, ``A survey on coverage path planning for
  robotics,'' \emph{Robotics and Autonomous Systems}, vol.~61, no.~12, pp.
  1258--1276, 2013.

\bibitem{ralphs2003capacitatedVRP}
T.~K. Ralphs, L.~Kopman, W.~R. Pulleyblank \emph{et~al.}, ``On the capacitated
  vehicle routing problem,'' \emph{Mathematical programming}, vol.~94, pp.
  343--359, 2003.

\bibitem{LI2011876}
Y.~Li, H.~Chen, M.~{Joo Er} \emph{et~al.}, ``Coverage path planning for uavs
  based on enhanced exact cellular decomposition method,'' \emph{Mechatronics},
  vol.~21, no.~5, pp. 876--885, 2011.

\bibitem{Hnidka_2019}
J.~Hnidka and D.~Rozehnal, ``Calculation of the maximum endurance of a small
  unmanned aerial vehicle in a hover,'' \emph{{IOP} Conference Series:
  Materials Science and Engineering}, vol. 664, no.~1, p. 012002, 2019.

\bibitem{Lussier2019enduranceSOTA}
M.~E. Lussier, J.~M. Bradley, and C.~Detweiler, \emph{Extending Endurance of
  Multicopters: The Current State-of-the-Art}.\hskip 1em plus 0.5em minus
  0.4em\relax AIAA, 2019.

\bibitem{mansouri_cooperative_2018}
S.~S. Mansouri, C.~Kanellakis, E.~Fresk \emph{et~al.}, ``Cooperative coverage
  path planning for visual inspection,'' \emph{Control Engineering Practice},
  vol.~74, pp. 118--131, 2018.

\bibitem{Kapoutsis2017DARP}
A.~C. Kapoutsis, S.~A. Chatzichristofis, and E.~B. Kosmatopoulos, ``Darp:
  Divide areas algorithm for optimal multi-robot coverage path planning,''
  \emph{Journal of Intelligent {\&} Robotic Systems}, vol.~86, no.~3, pp.
  663--680, 2017.

\bibitem{pathFinding}
P.~Surynek, ``Simple direct propositional encoding of cooperative path finding
  simplified yet more,'' in \emph{Nature-Inspired Computation and Machine
  Learning}, 11 2014, pp. 410--425.

\bibitem{9743616}
J.~Xie and J.~Chen, ``Multiregional coverage path planning for multiple energy
  constrained uavs,'' \emph{IEEE Transactions on Intelligent Transportation
  Systems}, vol.~23, no.~10, pp. 17\,366--17\,381, 2022.

\bibitem{10.1007/978-1-4471-1273-0_32}
H.~Choset and P.~Pignon, ``Coverage path planning: The boustrophedon cellular
  decomposition,'' in \emph{Field and Service Robotics}, A.~Zelinsky, Ed.\hskip
  1em plus 0.5em minus 0.4em\relax London: Springer London, 1998, pp. 203--209.

\bibitem{Guemri2016}
O.~Guemri, A.~Bekrar, B.~Beldjilali \emph{et~al.}, ``Grasp-based heuristic
  algorithm for the multi-product multi-vehicle inventory routing problem,''
  \emph{4OR}, vol.~14, no.~4, pp. 377--404, 2016.

\bibitem{8338417}
H.~Pham and Q.-C. Pham, ``A new approach to time-optimal path parameterization
  based on reachability analysis,'' \emph{IEEE Transactions on Robotics},
  vol.~34, no.~3, pp. 645--659, 2018.

\bibitem{mckinley1998cubic}
S.~McKinley and M.~Levine, ``Cubic spline interpolation,'' \emph{College of the
  Redwoods}, vol.~45, no.~1, pp. 1049--1060, 1998.

\bibitem{HertJINTHW_paper}
D.~Hert, T.~Baca, P.~Petracek \emph{et~al.}, ``{MRS} drone: A modular platform
  for real-world deployment of aerial multi-robot systems,'' \emph{Journal of
  Intelligent {\&} Robotic Systems}, vol. 108, no.~4, p.~64, Jul 2023.

\bibitem{Baca2021}
T.~Baca, M.~Petrlik, M.~Vrba \emph{et~al.}, ``{The MRS UAV System: Pushing the
  Frontiers of Reproducible Research, Real-world Deployment, and Education with
  Autonomous Unmanned Aerial Vehicles},'' \emph{Journal of Intelligent {\&}
  Robotic Systems}, vol. 102, no.~1, p.~26, 2021.

\bibitem{burri2015real-time}
M.~Burri, H.~Oleynikova,  \emph{et~al.}, ``Real-time visual-inertial mapping,
  re-localization and planning onboard mavs in unknown environments,'' in
  \emph{Intelligent Robots and Systems}, Sept 2015.

\bibitem{MRS2022ICUAS_HW}
D.~{Hert}, T.~{Baca}, P.~{Petracek} \emph{et~al.}, ``{MRS Modular UAV Hardware
  Platforms for Supporting Research in Real-World Outdoor and Indoor
  Environments},'' in \emph{International Conference on Unmanned Aircraft
  Systems}, 2022.

\bibitem{8594266}
T.~Baca, D.~Hert, G.~Loianno \emph{et~al.}, ``Model predictive trajectory
  tracking and collision avoidance for reliable outdoor deployment of unmanned
  aerial vehicles,'' in \emph{IEEE/RSJ International Conference on Intelligent
  Robots and Systems (IROS)}, 2018, pp. 6753--6760.

\bibitem{richter2016polynomial}
C.~Richter, A.~Bry, and N.~Roy, ``Polynomial trajectory planning for aggressive
  quadrotor flight in dense indoor environments,'' in \emph{Robotics Research},
  2016, pp. 649--666.

\end{thebibliography}



\end{document}